\documentclass{article}

\usepackage{microtype}
\usepackage{graphicx}
\usepackage{subfigure}
\usepackage{booktabs} 
\usepackage{multirow}
\usepackage{pifont}
\usepackage{verbatim}
\usepackage{booktabs} 
\usepackage{hyperref}
\usepackage{subfigure}
\usepackage{graphics}
\usepackage{amsfonts}
\usepackage{amsthm}
\usepackage[utf8]{inputenc}
\usepackage{multirow}
\usepackage{mathtools} 
\usepackage{makecell}
\usepackage{xspace}
\usepackage[table,xcdraw,dvipsnames]{xcolor}
\usepackage{enumitem}
\makeatletter
\renewcommand*\env@matrix[1][\arraystretch]{%
  \edef\arraystretch{#1}%
  \hskip -\arraycolsep
  \let\@ifnextchar\new@ifnextchar
  \array{*\c@MaxMatrixCols c}}
\makeatother

\usepackage{hyperref}

\usepackage[accepted]{icml2024}

\usepackage{amsmath}
\usepackage{amssymb}
\usepackage{mathtools}
\usepackage{amsthm}

\usepackage[capitalize,noabbrev]{cleveref}

\theoremstyle{plain}

\theoremstyle{definition}

\theoremstyle{remark}

\usepackage[textsize=tiny]{todonotes}

\newcommand{\mytitle}{Diffusion-based Episodes Augmentation \\ for Offline Multi-Agent Reinforcement Learning}
\newcommand{\myrunningtitle}{Diffusion-based Episodes Augmentation for Offline Multi-Agent Reinforcement Learning}
\newcommand{\alg}{\textbf{\texttt{EAQ}}\xspace}

\icmltitlerunning{\myrunningtitle}

\begin{document}

\twocolumn[
\icmltitle{\mytitle}




\begin{icmlauthorlist}
\icmlauthor{Jihwan Oh}{ai}
\icmlauthor{Sungnyun Kim}{ai}
\icmlauthor{Gahee Kim}{ai}
\icmlauthor{Sunghwan Kim}{ai}
\icmlauthor{Se-Young Yun}{ai}
\end{icmlauthorlist}

\icmlaffiliation{ai}{Kim Jaechul Graduate School of Artificial Intelligence, KAIST, Seoul, South Korea}

\icmlcorrespondingauthor{Se-Young Yun}
{yunseyoung@kaist.ac.kr}

\icmlkeywords{Machine Learning, ICML}

\vskip 0.3in
]



\printAffiliationsAndNotice{}  

\begin{abstract}
Offline multi-agent reinforcement learning (MARL) is increasingly recognized as crucial for effectively deploying RL algorithms in environments where real-time interaction is impractical, risky, or costly. In the offline setting, learning from a static dataset of past interactions allows for the development of robust and safe policies without the need for live data collection, which can be fraught with challenges. Building on this foundational importance, we present \alg, \textbf{E}pisodes \textbf{A}ugmentation guided by \textbf{Q}-total loss, a novel approach for offline MARL framework utilizing diffusion models. \alg integrates the Q-total function directly into the diffusion model as a guidance to maximize the global returns in an episode, eliminating the need for separate training. Our focus primarily lies on cooperative scenarios, where agents are required to act collectively towards achieving a shared goal—essentially, maximizing global returns. Consequently, we demonstrate that our episodes augmentation in a collaborative manner significantly boosts offline MARL algorithm compared to the original dataset, improving the normalized return by +17.3\% and +12.9\% for \textit{medium} and \textit{poor} behavioral policies in SMAC simulator, respectively.
\end{abstract}


\section{Introduction}
\label{Introduction}

Offline multi-agent reinforcement learning (MARL) \cite{levine2020offline} tackles the difficulties of implementing learning algorithms in situations where immediate interaction is infeasible or carries substantial dangers, such as operating in safety-critical systems like autonomous driving and robotic surgery. This approach, which emphasizes the utilization of a collection of past interaction data, presents a promising alternative to conventional MARL methods that rely on real-time data. By relying on pre-existing datasets, offline MARL enables the development of robust decision-making strategies without the need to continuously gather new data in potentially hazardous situations.

\begin{figure}[!t]
\centering
\includegraphics[width=\linewidth]{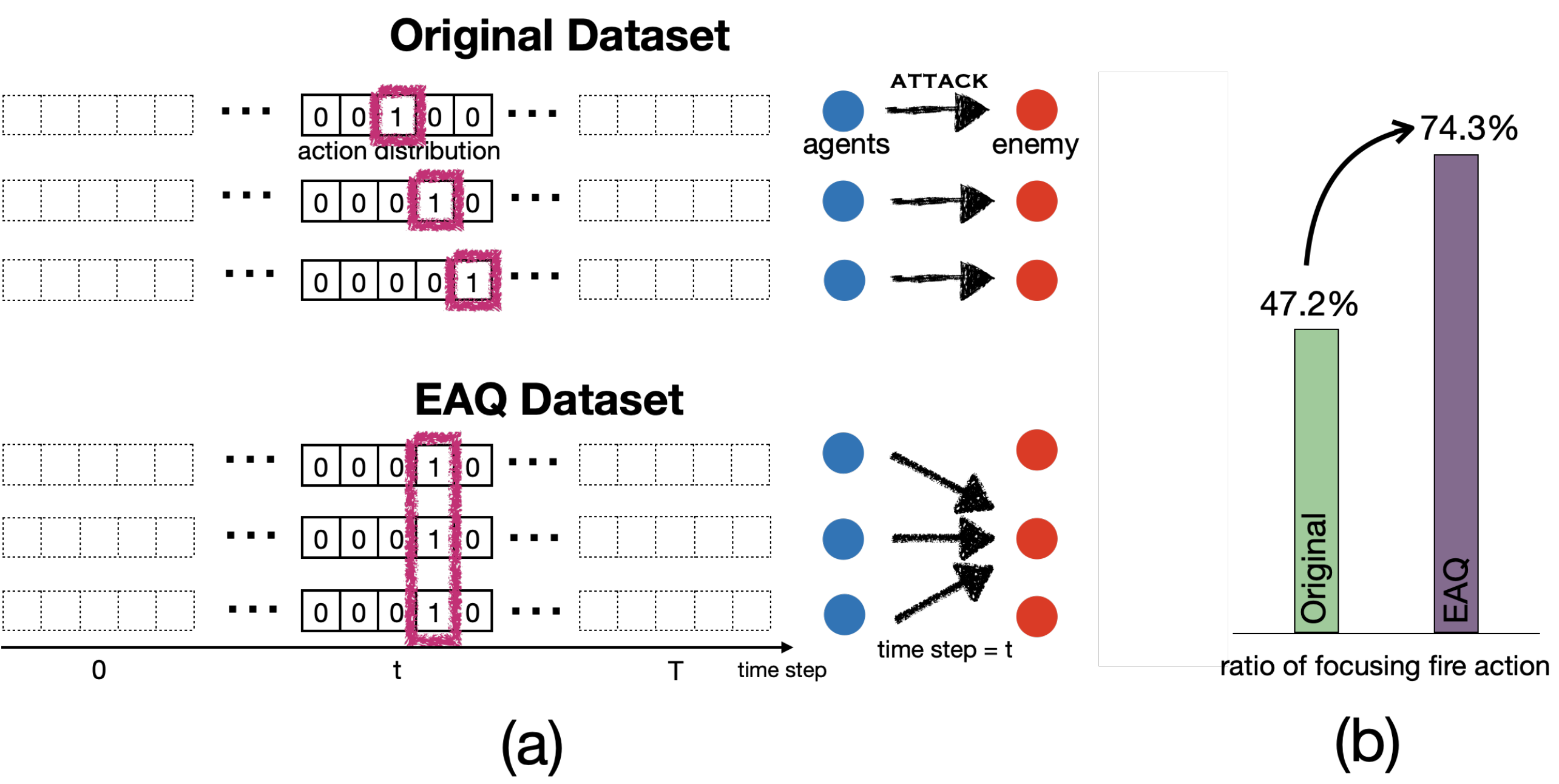}
\vspace{-20pt}
\caption{(a) \alg enhances the dataset to involve cooperative behaviors among agents. For instance, in a scenario like \textit{3marines vs 3marines} in StarCraft2, it encourages agents to sequentially target and attack a single enemy, which is a good strategy to win. (b) illustrates the proportion of focusing fire actions on a single enemy when all agents choose to attack. Compared to the original dataset, \alg has successfully increased the frequency of cooperative behaviors, towards achieving a shared goal.}
\label{fig:mot}
\end{figure}


However, collecting static data for offline MARL poses significant challenges, often requiring substantial time, expertise, and financial resources. This process typically involves human experts to check the relevance and accuracy of data, which not only makes it costly but also time-consuming. Hence, such static datasets usually lack the variability and complexity of real-world scenarios, resulting in a limited range of experiences for training MARL algorithms. This may lead to overfitting to the training data and poor generalization in actual environments, ultimately stifling the learning process and restricting the potential of MARL algorithms to adapt effectively to real-world conditions.

In order to tackle these problems, \citet{laskin2020reinforcement} and \citet{sinha2022s4rl} have proposed the use of data augmentation techniques in RL, which include introducing random variables into the original state space. The denoising diffusion probablistic model, proposed by \citet{ho2020denoising}, recommends the use of data augmentation in single-agent RL and has been supported by the studies of \citet{lu2024synthetic, he2024diffusion}. Yet, data augmentation strategies for MARL remain underexplored compared to the field of single-agent RL, although addressing data scarcity and efficiency is particularly crucial in offline MARL given its complex environments and interactions between multiple agents. In this paper, we introduce a novel episodes augmentation method, named as \alg, simplifying the training architecture while significantly enhancing performance through three primary contributions:



\vspace{-3pt}
\paragraph{Pioneering episodes augmentation for MARL:} We compile an offline dataset by incorporating multi-agent trajectories, including observations, actions, global rewards, and terminal states, formatted for a \texttt{Conv1D}-based diffusion model. As detailed in Section\,\ref{sec:method}, we describe our approach for effectively synthesizing episodes. To the best of our knowledge, our work is the first to propose a method for augmenting episodes specifically within the context of offline MARL. By identifying and addressing the need for enriched training scenarios, this study opens new avenues for data-centric research and applications in MARL.

\vspace{-3pt}
\paragraph{Cooperative trajectories augmentation:} Based on the prepared data, we guide our diffusion models to augment trajectories to be cooperative, \textit{i.e.}, exhibiting higher state-action values, with Q-total loss. This targeted approach ensures that the augmented trajectories are not only diverse but also strategically cooperative, enhancing the learning efficacy and performance of MARL algorithms in complex scenarios. Our findings indicate that our augmentation method, \alg, promotes increased cooperative behavior among agents. Figure \ref{fig:mot} depicts the SMAC \cite{samvelyan2019starcraft} scenario (\textit{3m vs 3m}), where we observe that \alg exhibits a significant rise in the cooperative actions among agents---represented as all agents attacking a single enemy to achieve maximum global returns.

\vspace{-3pt}
\paragraph{Simplification of model architecture:}
Q-total loss guidance \cite{wang2023diffusion, Ada_2024, kang2024efficient, chen2024boosting, li2023beyond} directly optimizes the diffusion model to maximize the expected cumulative rewards. This ensures that the diffusion model aligns closely with the objective function of RL, which is to learn from the reward function effectively. However, guiding with Q-total loss also requires a Q-function estimator $Q^{\text{tot}}_{\theta}(\mathbf{o}_{t}, \mathbf{a}_{t})$ to be trained separately \cite{wang2023diffusion, Ada_2024, kang2024efficient, chen2024boosting, li2023beyond}. In contrast, our method utilizes only a single diffusion model for episodes augmentation, eliminating the need for estimating Q-function. This not only simplifies the overall model architecture but also reduces the computational overhead.

\section{Preliminaries}

\subsection{Offline multi-agent reinforcement learning}
Offline multi-agent reinforcement learning (offline MARL) represents an integration of the unique dynamics and complexities of MARL with distinctive methods and constraints inherent in the offline RL. This integration entails a range of challenges and leverages combined techniques from both domains to address the intricacies of learning and decision-making in multi-agent environments without the requirement of real-time data collection.

Offline MARL is a Decentralized-Partially Observable Markov Decision Process (Dec-POMDP). The framework of Dec-POMDP is described by a tuple \( G := <S, A, P, r, \Omega, O, N, \gamma> \). In this context, \( s \in S \) denotes all possible environmental states and configurations for the system. \( a \in A \) represents the agents' actions, which may change the environment with specific probability. The transition function, $P$($s'$ $\mid$ $s$, $a$): $S \times A \times S'$ $\to$ [0, 1], determines the probability distribution of which state follows the current state after an agent's action. The reward function \( r \) measures the immediate payoff after an action in a specific state, assessing action's effectiveness. Furthermore, \( \Omega \) is the observation function which satisfy \(\Omega(s, a): S \times A \rightarrow O \) where $O$ is the observation space. \( N \) represents the number of environmental agents. The discount factor \( \gamma \) determines the importance of future rewards over immediate ones, affecting the agents' long-term strategies. Each agent acts based on its own observations and the shared goal of maximizing cumulative sum of rewards $\sum_{t=0}^{T}\gamma^{t} r_{t}$. To express the collective observations and actions from all agents at time step \( t \), we utilize bold notation $\mathbf{o}_t$ and \( \mathbf{a}_t \).

In our setup, agents are trained with a fixed dataset $\mathcal{D}:=\{\tau^{1}, \tau^{2}, ..., \tau^{E}\}$, where the trajectory $\tau$ is composed of sequential transitions $(\mathbf{o}_{t}, \mathbf{a}_{t}, r_{t}, done)$ where each refers to observation, action, and reward at timestep \textit{t} from multiple agents and $E$ is the number of trajectories. Generally, the policy is trained by maximizing the expectation of cumulative discounted global reward $\mathbb{E}\big[\sum^{T}_{t=0}\gamma^{t}r(\mathbf{o}_{t}, \mathbf{a}_{t})\big]$ \citep{sutton2018reinforcement}. Challenges specific to offline MARL dataset include (1) handling data from a scarce dataset, which cannot cover all the state-action space that increases exponentially as the number of agents grows, (2) mitigating non-stationarity in the environment due to the evolving strategies of other agents, and (3) deriving insights from possibly sub-optimal and exploratory past actions of these agents.


\subsection{Diffusion models}

Diffusion models \cite{ho2020denoising, song2020denoising} are generative models known for producing high-quality samples especially in vision domain. Recently, several works have applied diffusion models to RL tasks such as policy learning \cite{wang2023diffusion, kang2024efficient, Ada_2024, chen2023offline, li2023conservatism}, planning \cite{janner2022diffuser, liang2023adaptdiffuser}, and data synthesizing \cite{lu2024synthetic, he2024diffusion, li2023efficient, ni2023metadiffuser} utilizing their capabilities to capture the data distribution and powerful expressiveness. 

Diffusion models operate through two main phases: the forward noising process and the reverse denoising process. In the forward phase, data samples \( \mathbf{x}_0 \) from the original distribution $q(\mathbf{x}_{0})$ are progressively transformed into Gaussian noise $\mathcal{N}(\mathbf{0}, \mathbf{I})$ over a series of diffusion steps \( k \) by adding i.i.d. Gaussian noise with a standard deviation $\sigma$. The transformation at each step according to a variance schedule $\beta_{1}, ..., \beta_{K}$ is given by:
\begin{equation}
q(\mathbf{x}_{1:K}|\mathbf{x}_{0}) := \prod_{k=1}^K q(\mathbf{x}_{k}|\mathbf{x}_{k-1}) 
\end{equation}

In the reverse phase, the model learns to reconstruct the original data from the noised starting point $p(\mathbf{x}_{K}) = \mathcal{N}(\mathbf{x}_K;\mathbf{0}, \mathbf{I})$. This process is modeled by:
\begin{equation}
p_{\theta}(\mathbf{x}_{k-1} | \mathbf{x}_k) := \mathcal{N}(\mathbf{x}_{k-1}; \boldsymbol{\mu}_{\theta}(\mathbf{x}_k, k), \mathbf{\Sigma}_{\theta}(\mathbf{x}_k, k)),
\end{equation}

where \( \boldsymbol{\mu}_{\theta}(\mathbf{x}_k, k) \) and \( \boldsymbol{\Sigma}_{\theta}(\mathbf{x}_k, k) \) are functions parameterized by neural networks, representing the mean and covariance of the Gaussian distribution at each reverse step. Finally, training is performed via optimizing the variational lower bound on negative log likelihood:
\begin{equation}
\mathbb{E}[\mathrm{log}p_{\theta}(\mathbf{x}_{o})] \ge \mathbb{E}_{q}[\mathrm{log}\frac{p_{\theta}{(\mathbf{x}_{0:T})}}{q(\mathbf{x}_{1:T}|\mathbf{x}_{0})}]    
\end{equation}
In this paper, we denote trajectory $\tau_{k}$ as $\sqrt{\bar{\alpha}_{k}}\tau_{0} + \sqrt{1-\bar{\alpha}_{k}}\epsilon$ where $\alpha_{k} := 1- \beta_{k}$ and $\bar{\alpha}_{k} := \prod_{s=1}^k \alpha_{s}$, where $\tau_{0}$ refers the trajectory which is not corrupted with gaussian noise. By leveraging diffusion models to planning domain, \citet{zhu2023madiff, li2023beyond} could significantly improve their performance with MARL algorithms.




\subsection{Data synthesis in RL}
Conventional data augmentation methods in RL typically introduce minor perturbations to states to maintain consistency with the environment dynamics \cite{laskin2020reinforcement, sinha2022s4rl}. However, diffusion models, commonly used in computer vision to create synthetic data, are well-suited for addressing data scarcity in the RL datasets and offer a more robust approach by learning the entire distribution from the actual dataset, \(\mathcal{D}_{\text{real}}\). \citet{lu2024synthetic} first parameterize and learn the data distribution \(\rho_{\theta}(\tau)\) from \(\mathcal{D}_{\text{real}}\), and then generate the desired synthetic data, represented as \(\mathcal{D}_{\text{syn}} = \{\tau \sim \rho_{\theta}(\tau)\}\). The resulting composite dataset, \(\mathcal{D} = \mathcal{D}_{\text{real}} \cup \mathcal{D}_{\text{syn}}\), is then utilized for policy learning. 

In online settings, ongoing interaction between agent and environment facilitates a cyclical updating mechanism where both the diffusion model and the policy itself are iteratively refined. Such a dynamic process promotes continual enhancements in the policy's decision-making capabilities and the diffusion model's data generation accuracy. The ability to adapt and improve continuously is crucial for maintaining relevance and effectiveness in changing environments.

Data augmentation strategies for MARL are not as advanced as those developed for single-agent RL. Addressing this discrepancy is particularly essential given the complex and multifaceted dynamics inherent in MARL systems, where multiple agents interact within a shared space. These interactions can significantly complicate the training process, making the development of robust data augmentation techniques vital. In offline settings, where real-time data collection is not feasible, the scarcity of diverse and comprehensive datasets becomes a critical issue. Effective data augmentation for MARL is necessitated to simulate a wider range of scenarios and interactions, enhancing the model's ability to generalize from limited data. 

\section{Synthesizing Multi-Agent Offline Data}
\label{sec:method}

In the realm of synthesizing data for RL, methodologies can generally be categorized into two primary classes: (1) Purturbation on state space \cite{laskin2020reinforcement, sinha2022s4rl} and (2) Generative models \cite{lu2024synthetic, he2024diffusion}. Our approach falls within the generative class, which focuses on creating new data points through models that learn the underlying distribution of the dataset. For that reason, generative models are more advantageous to highlight the diversity of the data rather than merely increasing the quantity by making the data similar.

One of the key challenges in MARL data augmentation is the complexity associated with handling data from multiple agents. The difficulties primarily arise from two main areas: (1) integration of multi-agent data which is often unclear due to the intricate relationships and dependencies between agents' actions and states; and (2) development of methods to enhance the cooperativeness of episodes, which is crucial for scenarios where agents need to work together towards a common goal.
To address these challenges, we devise specific strategies as outlined below.

\paragraph{Integration of agents’ features.} We have established a method to efficiently merge attributes from numerous agents, guaranteeing that the data accurately represents the interactions between the agents. This entails employing sophisticated aggregation techniques that preserve the integrity and context of every agent's actions and observations, enabling a comprehensive perspective of the multi-agent environment.
   
\paragraph{Enhancing episode cooperativeness.} We focus on generating the episodes to foster more cooperative interactions among agents. This is achieved through the use of Q-total loss directly into the diffusion loss. By adpoting the cumulative sum of reward given all agents' current observation and action, generally denoted as $Q^{\text{tot}}(\mathbf{o}_{t}, \mathbf{a}_{t})$ in RL domain, we promote the generative model to synthesize the episodes to be more cooperative direction, which is key to improving the overall effectiveness of the training episodes.


In the rest of this section, we present how we devised novel methods for synthesizing pre-existing MARL data, with the goal of improving the quality of offline datasets. Our method use a \texttt{Conv1D}-based diffusion model to create the fundamental elements of MARL episodes $\tau$: observations $\mathbf{o}_{t}$, actions $\mathbf{a}_{t}$, rewards $r_{t}$, terminal states $done_{t}$, and total Q-values $Q^\text{tot}(\mathbf{o}_{t}, \mathbf{a}_{t})$ (we will denote this as $Q_{t}^\text{tot}$ for simplicity) which refers the sum of cumulative rewards given the current state and action until the episode ends, named as \textit{rewards-to-go} in RL domain. We augment the episodes to be more cooperative by guiding with $Q_{t}^\text{tot}$.



We clarify our analytical approach and its implementation in section \ref{sec:training_data_pre} and \ref{sec:loss_guidance}, illustrating how these strategies are used with MARL data. Furthermore, we present empirical evidence of substantial performance improvement when employing these enhanced datasets with offline MARL methods, as demonstrated in Table \ref{table:main}.

\begin{figure*}[t]
\vspace{10pt}
\centering
\small
\centerline{\includegraphics[width=2.1\columnwidth]{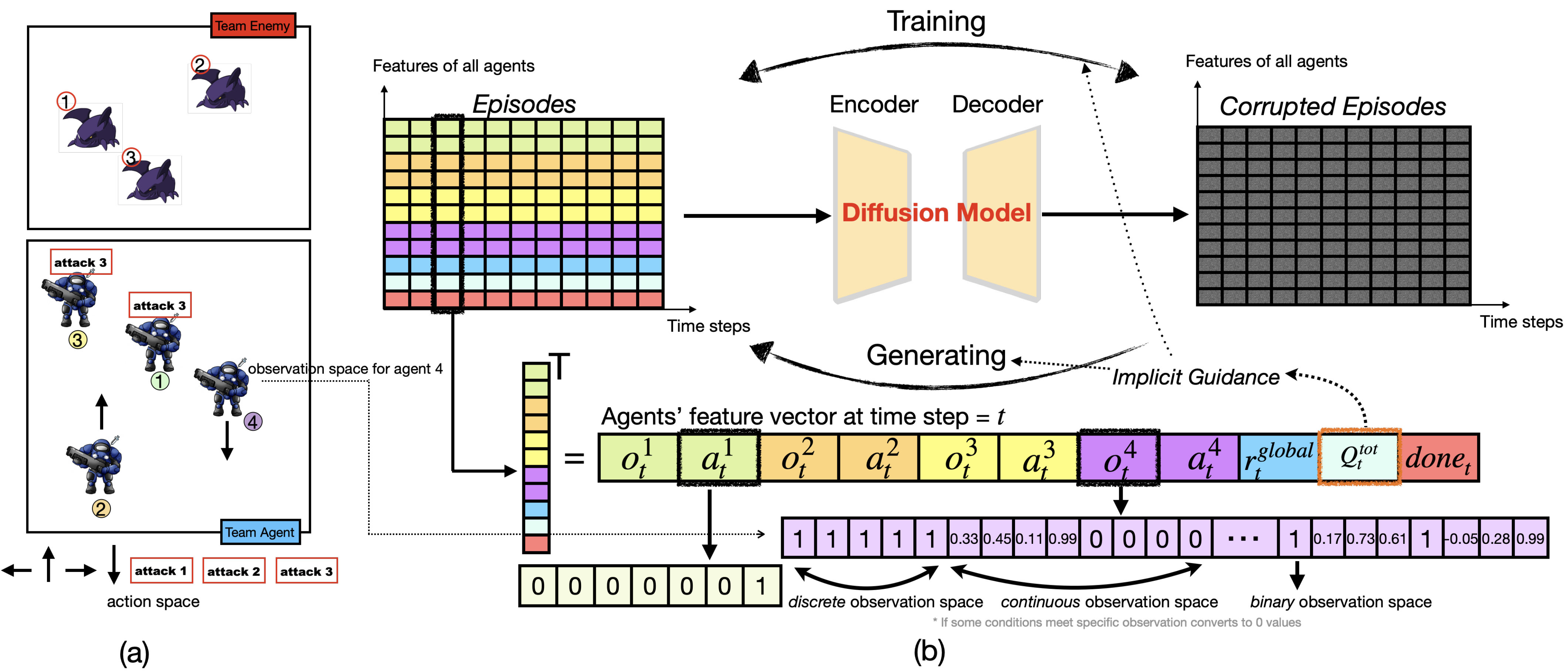}}
\vspace{-10pt}
\caption{Our proposed algorithm architecture for MARL episodes augmentation. (a) represents the example of MARL offline dataset based on SMAC \cite{samvelyan2019starcraft} environments at time step = $t$. There are two teams that have to kill each other to win the battle. Here, the actions are in the discrete space such as digit which is converted into one-hot encoding format. (b) represents (1) how our training datasets are prepared and look like, (2) how \alg is trained under the guidance of $Q_{t}^{\text{tot}}$ and (3) how highly the values in the datasets is  multi-modal.}
\label{fig:main}
\end{figure*}

\subsection{Training data preparation}
\label{sec:training_data_pre}
In this section, we describe how we prepare training data, which involves reformatting the original data into a specific structure for diffusion model with \texttt{Conv1D}. The data structure is reformatted as the size of (\textbf{B}, \textbf{F}, \textbf{T}), serving as the input format for our model: \textbf{B}atch size refers to the number of episodes included in the given dataset $\mathcal{D}$, \textbf{F}eature length represents the combined length of the feature vectors for all agents, and \textbf{T}ime steps indicate the maximum duration of any episode within the dataset, $\mathcal{D}$. If the episode ends before the maximum time step in this dataset, then its feature values are padded with 0. Our dataset exhibits high multi-modality, consisting of diverse range of features from multiple agents that include both continuous and discrete values, as shown in Figure \ref{fig:main}.

To address the discrete action space, we convert discrete actions into a one-hot encoding format, as depicted in Figure \ref{fig:main}. In the last few dimensions of the feature space, we append several critical pieces of information:
\vspace{-0.1in}
\begin{itemize}[leftmargin=*]
\setlength\itemsep{0pt}
    \item \textbf{Global reward} \(r^\text{global}_{t}\) given to the agents at time \(t\).
    \item \textbf{Total state-action value} $Q^\text{tot}(\mathbf{o}_t, \mathbf{a}_t)$ evaluating the chosen actions' value given the states, particularly used for guiding generated episodes to be cooperative.
    \item \textbf{Episode termination state} indicating whether the episode terminates at the current step. We mark it with 1 if the episodes ends, or 0 otherwise.
\end{itemize}
\vspace{-0.1in}
Then, our episode data reformatted for training is shaped as:

\begin{equation}
\tau_{0} :=
\begin{bmatrix}[1.3]
o_{0}^{1} & o_{1}^{1} & \cdots & o_{T}^{1} \\
\vspace{0.1in}
a_{0}^{1} & a_{1}^{1} & \cdots & a_{T}^{1}\\
\vdots & \ddots & \ddots & \vdots \\
o_{0}^{N} & o_{1}^{N} & \cdots & o_{T}^{N} \\
a_{0}^{N} & a_{1}^{N} & \cdots & a_{T}^{N}\\
r_{0}^{global} & r_{1}^{global} & \cdots & r_{T}^{global}\\
Q_{0}^{\text{tot}} & Q_{1}^{\text{tot}} & \cdots & Q_{T}^{\text{tot}}\\
done_{0} & done_{1} & \cdots & done_{T}\\
\end{bmatrix},
\end{equation}
where the superscript and subscript of observation $o_{t}^{n}$ and action $a_{t}^{n}$ denote the agent's index and the time step of the episode. The details on the each value is deployed in the Figure \ref{fig:main}. This approach allows our model to perform observation learning, policy learning, reward learning, and even state-action value learning. We use only a diffusion model for these tasks, which significantly reduces the complexity of training and sampling processes. This streamlined approach ensures our model to efficiently learn the pre-collected trajectories distribution, while guiding its episodes synthesis in a cooperative manner. The detailed specifications for the dataset $\mathcal{D}$ is presented in Section\,\ref{subsec:dataset}.

\subsection{Implicit $Q^\text{tot}$ loss guidance augmentation}
\label{sec:loss_guidance}

In training session, we train $Q^\text{tot}({\textbf{o}_{t}, \textbf{a}_{t}})$ whose learning target is sum of reward given the current state and action until the episode ends, named as \textit{reward-to-go} in RL domain, which can be calculated given the dataset. The reason why we utilize the \textit{reward to go} is that it indicates the unbiased estimator of the state-action value from the current timestep, defined by $Q(o_{t}, a_{t}) = \mathbb{E}[\Sigma_{t'=t}^{T}\gamma^{t'-t} r(o, a)|o=o_{t'}, a=a_{t'}]$. From this, we can implicitly train our diffusion model to make more cooperative scenarios by maximizing $Q_{f_{\theta}(\tau_{k}, k)}(\textbf{o}, \textbf{a})$, which can be represented as:
$\mathop{\arg \max}_{f_{\theta}} Q_{f_{\theta}(\tau_{k}, k)}(\textbf{o}, \textbf{a})$, where $f_{\theta}$ is diffusion model and $Q_{f_{\theta}(\tau_{k}, k)}(\textbf{o}, \textbf{a})$ is regarded as the estimator of $Q_{t}^{\text{tot}}$ generated from the diffusion model as shown in Figure \ref{fig:main} (b). 

In this context, we optimize the diffusion model with the loss $\mathbb{E}_{k\in U[1, T], \tau_{0}\in \mathcal{B} \sim \mathcal{D}}[\|\tau_{0} - f_{\theta}(\tau_{k}, k)\|^2]$ to predict the start trajectory $\tau_{0}$, not just like DDPM \cite{ho2020denoising} which predict the noise $\epsilon$ with the loss $\mathbb{E}_{k\in U[1, T], \tau_{0}\in \mathcal{B} \sim \mathcal{D}}[\|\epsilon - \epsilon_{\theta}(\tau_{k}, k)\|^2]$. These two formulations are interchangeable and are both widely used in diffusion models \cite{ramesh2022hierarchical}. Our diffusion loss function to predicting the start trajectory is for utilizing the $Q_{t}^{\text{tot}}$ as target values with the estimator of $Q_{f_{\theta}(\tau_{k}, k)}(\textbf{o}, \textbf{a})$ in the training phase, simultaneously maximizing the estimator. If we predict the noise $\epsilon_{\theta}$, then we cannot optimize the $Q_{f_{\theta}(\tau_{k}, k)}(\textbf{o}, \textbf{a})$ to be maximized in the training steps otherwise, it will require additional model to predict the $Q^{\text{tot}}$. Then, the final loss function with diffusion model is given by:
\begin{align}
\mathcal{L}(\theta) =&~\mathcal{L}_\text{diffusion} + \lambda \mathcal{L}_{Q_\text{tot}} \nonumber \\ 
                    =&~\mathbb{E}_{k\in U[1, K], \tau_{0}\in \mathcal{B} \sim \mathcal{D}}[\|\tau_{0} - f_{\theta}(\tau_{k}, k)\|^2 \\
                     &+ \lambda  \left[ \max_{\tau_{0} \in \mathcal{B}} Q_{\tau_{0}}(\textbf{o}, \textbf{a})-Q_{f_{\theta}(\tau_{k}, k)}(\textbf{o}, \textbf{a}) \right]_{\geq 0}] \nonumber
\end{align}


where $\max_{\tau_{0} \in \mathcal{B}} Q_{\tau_{0}}(\textbf{o}, \textbf{a})$ is the maximum Q-value empirically obtained from the minibatch $\mathcal{B}$.
$\lambda$ is a balancing hyperparameter, and $Q_{\tau_{0}}(\textbf{o}, \textbf{a})$ is an expectation of $Q^\text{tot}_{t}$ in an episodes represented as $\frac{1}{T}\Sigma_{t=0}^{t=T} Q_{t}^{\text{tot}}$. We define the upper bound of $Q_{f_{\theta}(\tau_{k}, k)}(\textbf{o}, \textbf{a})$ to be $\max_{\tau_{0} \in \mathcal{B}} Q_{\tau_{0}}(\textbf{o}, \textbf{a})$ in the minibatch $\mathcal{B}$ for the stable training, otherwise the value generated by $Q_{f_{\theta}(\tau_{k}, k)}(\textbf{o}, \textbf{a})$ can explode to the infinity value by the reason that we maximize the $Q_{f_{\theta}(\tau_{k}, k)}(\textbf{o}, \textbf{a})$. The logic to train \alg is explained in Algorithm \ref{alg:1}.

\begin{algorithm}[tb]
   \caption{\alg}
   \label{alg:1}
\begin{algorithmic}
   \STATE {\bfseries Given:} dataset $episodes_N  \in\mathcal{D}_{\text{real}}$
   \STATE {\bfseries Set:} hyperparameter $\lambda$
   \STATE Calculate the \textit{reward-to-go} in the episodes.
   \STATE Transform $\mathcal{D}$ into (B, F, T) shaped tensor. 
   \STATE \textcolor{gray}{\textit{/* Training Process */}}
   \REPEAT
   \STATE $\mathbf{x}_0 \sim q(\mathbf{x}_{0})$
   \STATE $k \sim \mathrm{Uniform}(\{1, ..., K\})$
   \STATE Compute the loss 
   \\$\mathcal{L}_{\text{diffusion}} = \|\tau_{0} - f_{\theta}(\tau_{k}, k)\|^2$
   \\$\mathcal{L}_{Q_\text{tot}} = \left[ \max_{\tau_{0} \in \mathcal{B}} Q_{\tau_{0}}(\textbf{o}, \textbf{a})-Q_{f_{\theta}(\tau_{k}, k)}(\textbf{o}, \textbf{a}) \right]_{\geq 0}$
   \\ $\mathcal{L}(\theta) = \mathbb{E}_{k\in U[1, K], \tau_{0}\in \mathcal{B} \sim \mathcal{D}}[\mathcal{L}_{\text{diffusion}} + \lambda\mathcal{L}_{Q_\text{tot}}]$
   \UNTIL{converged}
   
   \STATE \textcolor{gray}{\textit{/* Sampling Process */}}
    \STATE {\bfseries Set:} Upsampling scale $S$
   \STATE Generate \(episodes_{N \times S} \in \mathcal{D}_{\text{syn}}\)
   \STATE \( \mathcal{D}_{\text{aug}} = \mathcal{D}_{\text{real}} \cup \mathcal{D}_{\text{syn}}\)
\end{algorithmic}
\end{algorithm}


\subsection{Visualization of augmented dataset}

\begin{figure}[!h]
\centering
\includegraphics[width=\linewidth]{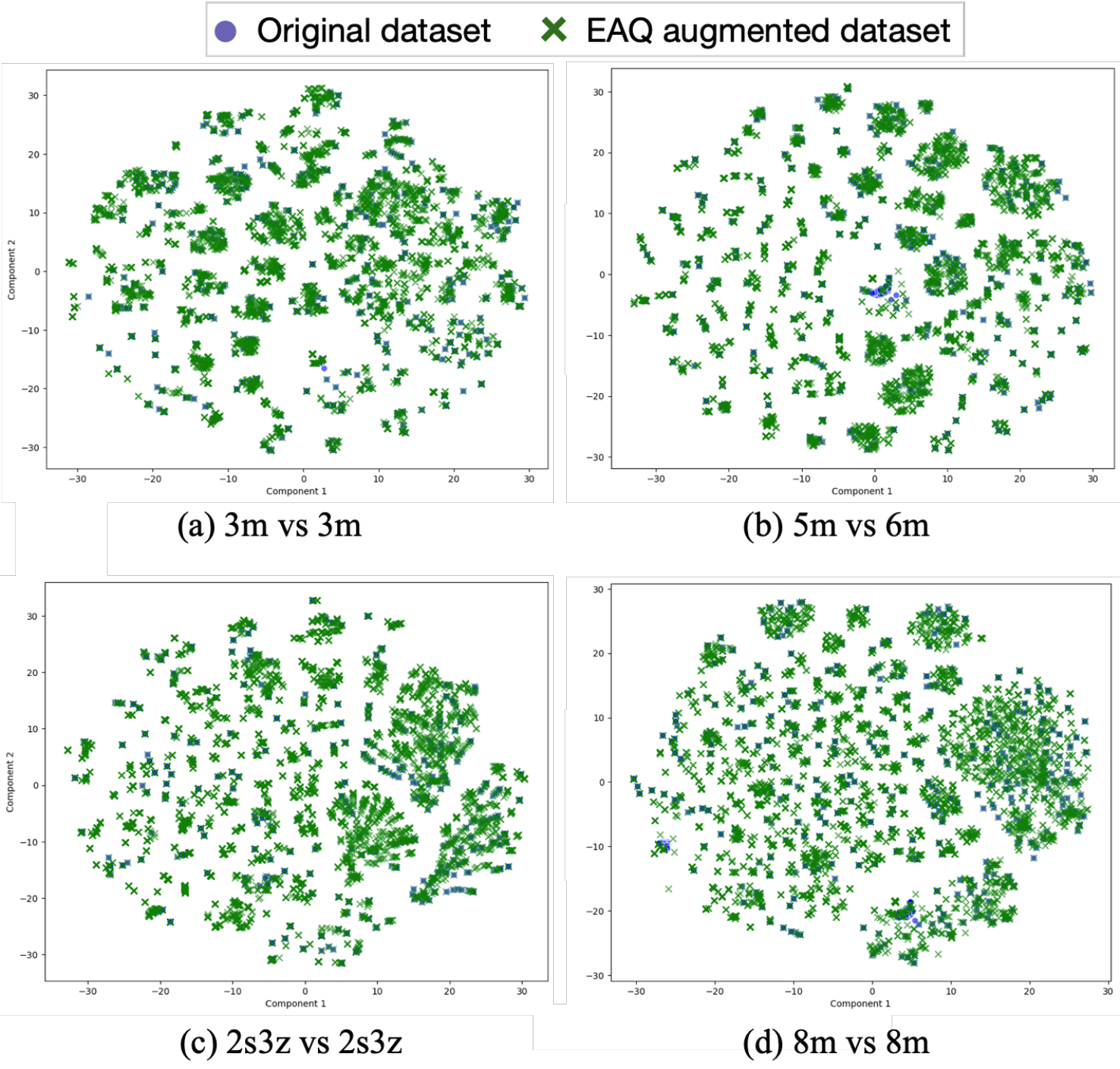}
\vspace{-15pt}
\caption{We visualize a t-SNE \cite{JMLR:v9:vandermaaten08a} projection on the observation space of original dataset and \alg augmented dataset on several scenarios in SMAC environment.}
\label{fig:t-sne}
\end{figure}

We wonder if \alg is indeed beneficial to augment the data in the perspective of variability and cooperativeness.
Thus, we raise a question: \textit{``Does \alg-generated dataset encompass a broader range of states than the original dataset?''}
The cooperativeness of augmented data by \alg has been demonstrated in Section \ref{Introduction} with Figure \ref{fig:mot}. To validate the variability of the augmented data, we map the observations of the agents into 2-D space using t-SNE \cite{JMLR:v9:vandermaaten08a} as shown in Figure \ref{fig:t-sne}. We find that \alg not only follows the true data distribution but also covers broader range of observations, which shows the fidelity and variability of \alg.

\section{Experiments}

\begin{table*}[t]
    \centering
\caption{Returns on each SMAC scenario using CQL and BCQ-based algorithms, reported with mean and standard deviation across 5 runs for 15,000 training iterations. Each scenario consists of \{\textit{team}\}\,-\,\{\textit{quality}\}, where m, s, and z in \textit{team} refer to marine, stalker, and zealot, respectively. The original data \textit{quality} has medium and poor policy. The best performance is marked with \textbf{bold}.}
    \label{table:main}
    \vspace{5pt}
    \addtolength{\tabcolsep}{-1pt}
     \resizebox{\textwidth}{!}{%
    \begin{tabular}{lccccc|ccccc}
    \toprule
    & \multicolumn{5}{c}{QMIX-CQL} & \multicolumn{5}{c}{QMIX-BCQ} \\
    \cmidrule(r){2-6} \cmidrule(r){7-11}
    Scenario & original & RAD-s & RAD-m & $\alg^{-Q}$ & \alg(ours) & original & RAD-s & RAD-m & $\alg^{-Q}$ & \alg(ours) \\
    \midrule
    \multirow{1}{*}{\makecell{3m-medium}}       & \multirow{1}{*}{\makecell{7.47 \scriptsize $\pm$ 6.35}} & \textbf{16.90} \scriptsize $\pm$ 2.75 & 15.05 \scriptsize $\pm$ 3.11 &  \multirow{1}{*}{\makecell{8.22 \scriptsize $\pm$ 5.77}} & \multirow{1}{*}{13.70 \scriptsize $\pm$ 7.43}  & 5.17 \scriptsize $\pm$ 2.76 & 2.56 \scriptsize $\pm$ 1.45 & 4.56 \scriptsize $\pm$ 2.00 & 4.16 \scriptsize $\pm$ 3.27 & \textbf{16.90} \scriptsize $\pm$ 2.21  \\
    \multirow{1}{*}{\makecell{3m-poor}}       & \multirow{1}{*}{\makecell{3.81 \scriptsize $\pm$ 1.59}} & 2.83 \scriptsize $\pm$ 0.85 & 7.87 \scriptsize $\pm$ 6.36  & \multirow{1}{*}{\makecell{6.95 \scriptsize $\pm$ 3.46}} & \multirow{1}{*}{\textbf{10.60} \scriptsize $\pm$ 6.25}    & 0.79 \scriptsize $\pm$ 0.83 & 1.35 \scriptsize $\pm$ 1.31 & 0.77 \scriptsize $\pm$ 0.32 & 1.38 \scriptsize $\pm$ 0.97  & \textbf{1.59} \scriptsize $\pm$ 1.59 \\
    \hline
    \multirow{1}{*}{\makecell{5m vs 6m-medium}}       & \multirow{1}{*}{\makecell{0.29 \scriptsize $\pm$ 0.45}} & 0.05 \scriptsize $\pm$ 0.07 & 0.18 \scriptsize $\pm$ 0.31  &  \multirow{1}{*}{\makecell{0.77 \scriptsize $\pm$ 0.42}} & \multirow{1}{*}{\textbf{3.75} \scriptsize $\pm$ 2.04}  & 3.38 \scriptsize $\pm$ 0.43 & 3.00 \scriptsize $\pm$ 0.81 & \textbf{3.52} \scriptsize $\pm$ 0.52 & 2.62 \scriptsize $\pm$ 1.87 & 1.72 \scriptsize $\pm$ 1.66  \\
    \multirow{1}{*}{\makecell{5m vs 6m-poor}}       & \multirow{1}{*}{\makecell{0.99 \scriptsize $\pm$ 1.21}}& 1.84 \scriptsize $\pm$ 1.83 & 1.73 \scriptsize $\pm$ 1.61  &  \multirow{1}{*}{\makecell{\textbf{6.48} \scriptsize $\pm$ 1.39}} & \multirow{1}{*}{3.64 \scriptsize $\pm$ 2.24}    & 2.01 \scriptsize $\pm$ 1.05 & 1.96 \scriptsize $\pm$ 1.27 & 2.25 \scriptsize $\pm$ 1.38 & 4.00 \scriptsize $\pm$ 1.53 & \textbf{4.23} \scriptsize $\pm$ 0.70 \\
    \hline
    \multirow{1}{*}{\makecell{8m-medium}}       & \multirow{1}{*}{\makecell{1.87 \scriptsize $\pm$ 0.86}} & 1.24 \scriptsize $\pm$ 0.67 & 1.22 \scriptsize $\pm$ 0.70 &  \multirow{1}{*}{\makecell{2.95 \scriptsize $\pm$ 1.17}} & \multirow{1}{*}{\textbf{3.59} \scriptsize $\pm$ 0.60}  & 1.64 \scriptsize $\pm$ 1.10 & 2.32 \scriptsize $\pm$ 0.75 & 1.60 \scriptsize $\pm$ 1.83 & 4.91 \scriptsize $\pm$ 1.34 & \textbf{5.92} \scriptsize $\pm$ 5.58  \\
    \multirow{1}{*}{\makecell{8m-poor}}       & \multirow{1}{*}{\makecell{0.35} \scriptsize $\pm$ 0.25} & 0.66 \scriptsize $\pm$ 0.62 & 0.88 \scriptsize $\pm$ 0.92 &  \multirow{1}{*}{\makecell{1.41} \scriptsize $\pm$ 0.93} & \multirow{1}{*}{\textbf{1.69} \scriptsize $\pm$ 1.69} & 2.56 \scriptsize $\pm$ 0.48 & 2.75 \scriptsize $\pm$ 0.42 & 2.44 \scriptsize $\pm$ 0.41 & \textbf{9.23} \scriptsize $\pm$ 2.61  & 6.06 \scriptsize $\pm$ 1.79 \\
    \hline
    \multirow{1}{*}{\makecell{2s3z-medium}}       & \multirow{1}{*}{\makecell{3.77 \scriptsize $\pm$ 0.86}} & 5.98 \scriptsize $\pm$ 2.92 & 5.26 \scriptsize $\pm$ 1.48 &  \multirow{1}{*}{\makecell{\textbf{9.74} \scriptsize $\pm$ 2.15}} & \multirow{1}{*}{9.05 \scriptsize $\pm$ 1.80}  & 6.86 \scriptsize $\pm$ 0.84 & 5.41 \scriptsize $\pm$ 1.78 & 6.57 \scriptsize $\pm$ 1.49 & 5.65 \scriptsize $\pm$ 1.68 & \textbf{7.46} \scriptsize $\pm$ 2.61  \\
    \multirow{1}{*}{\makecell{2s3z-poor}}       & \multirow{1}{*}{\makecell{5.16} \scriptsize $\pm$ 2.71} & 5.13 \scriptsize $\pm$ 1.68 & 3.30 \scriptsize $\pm$ 1.43 &  \multirow{1}{*}{\makecell{\textbf{6.13} \scriptsize $\pm$ 1.26}} & \multirow{1}{*}{5.40 \scriptsize $\pm$ 3.37}    & 5.20 \scriptsize $\pm$ 1.89 & 5.39 \scriptsize $\pm$ 1.33 & 4.72 \scriptsize $\pm$ 0.87 & 6.61 \scriptsize $\pm$ 1.38  & \textbf{7.34} \scriptsize $\pm$ 3.90 \\
    \hline
    \multirow{1}{*}{\makecell{3s5z vs 3s6z-medium}}       & \multirow{1}{*}{\makecell{2.69 \scriptsize $\pm$ 1.49}} & 2.81 \scriptsize $\pm$ 1.00 & 3.34 \scriptsize $\pm$ 1.37 &  \multirow{1}{*}{\makecell{6.14 \scriptsize $\pm$ 2.60}} & \multirow{1}{*}{\textbf{7.17} \scriptsize $\pm$ 1.37}  & 6.14 \scriptsize $\pm$ 1.86 & 5.37 \scriptsize $\pm$ 1.21 & 5.47 \scriptsize $\pm$ 0.79 & 6.07 \scriptsize $\pm$ 1.22 & \textbf{6.58} \scriptsize $\pm$ 2.90  \\
    \multirow{1}{*}{\makecell{3s5z vs 3s6z-poor}}       & \multirow{1}{*}{\makecell{2.87 \scriptsize $\pm$ 0.80}} & 2.69 \scriptsize $\pm$ 0.42 & 2.00 \scriptsize $\pm$ 0.88 &  \multirow{1}{*}{\makecell{5.16 \scriptsize $\pm$ 1.19}} & \multirow{1}{*}{\textbf{5.35} \scriptsize $\pm$ 0.74}    & 4.69 \scriptsize $\pm$ 1.32  & 5.48 \scriptsize $\pm$ 1.23 & 5.75 \scriptsize $\pm$ 0.40 & 5.51 \scriptsize $\pm$ 0.37 & \textbf{6.37} \scriptsize$\pm$ 0.46 \\
    \hline
    \midrule
    \multirow{1}{*}{\makecell{AVG-medium}}       & \multirow{1}{*}{\makecell{3.22}} & 5.39 & 5.01 &  \multirow{1}{*}{\makecell{5.56}} & \multirow{1}{*}{\textbf{7.45}}  & 4.64 & 3.73 & 4.34 & 4.68 & \textbf{7.72} \\
    \multirow{1}{*}{\makecell{AVG-poor}}       & \multirow{1}{*}{\makecell{2.64}} & 2.63 & 3.16 &  \multirow{1}{*}{\makecell{5.23}} & \multirow{1}{*}{\textbf{5.34}}    & 3.05  & 3.38 & 3.19 & \textbf{5.35} & 5.12 \\
    \bottomrule
    \end{tabular}
     }%
\end{table*}

We validate \alg practically with other RL data augmentation methods based on the offline MARL algorithms. For the experiments, we consider two variants of offline MARL algorithms adopted to QMIX \cite{rashid2018qmix} with representatives of the offline RL baseline: QMIX-CQL \cite{kumar2020conservative} and QMIX-BCQ \cite{fujimoto2019offpolicy}. We upsample the dataset five times more than the original size and aggregate it as \( \mathcal{D}_{\text{aug}} = \mathcal{D}_{\text{real}} \cup \mathcal{D}_{\text{syn}}\). 


\subsection{Baseline RL algorithms}
As aforementioned, we utilize the combined version of QMIX with CQL and BCQ algorithms implemented by \citet{formanek2023offthegrid}. We briefly explain about each algorithm below.

\textbf{QMIX} \cite{rashid2018qmix} addresses the challenge of learning decentralized policies in a centralized end-to-end fashion, leveraging global state information during training. It employs a mixing network that estimates the joint action-value function $Q^{\text{tot}}_{t}$ as a monotonic combination of per-agent utility functions $Q^{agent}_{t}$. 

\textbf{CQL} \cite{kumar2020conservative} aims to penalize the Q-values of out-of-distribution actions in the offline data, ensuring that the Q-function underestimates rather than overestimates. It learns a Q-function network by minimizing the standard Bellman error, with an additional regularization term that penalizes Q-values for actions with low probability under the data distribution from the offline dataset.

\textbf{BCQ} \cite{fujimoto2019offpolicy} is an offline reinforcement learning algorithm that mitigates distributional shift by constraining the policy to the behavior distribution. It starts by training a generative model, such as a VAE, on the offline dataset to capture the action distribution of the behavior policy conditioned on the states. For a given observation $o_{t}$, BCQ samples multiple actions from the generative model's, denoted as $G_{\theta}$, output distribution $ G_{\theta}(\cdot|o_{t}) $. It then selects the action $ a^*_{t} $ that maximizes the Q-value $ Q(o_{t}, a_{t}) $ while remaining close to the behavior distribution $ G_{\theta}(a_{t}|o_{t}) $. 

\subsection{Compared augmentation algorithms}
Augmentation in RL remains largely unexplored, particularly in proprioceptive observation-based single-agent RL, and even more so in the multi-agent RL domain. For the purposes of comparison with existing augmentation methods, we utilize RAD \cite{laskin2020reinforcement} as our benchmark model. Furthermore, to assess the effectiveness of the \(Q^{\text{tot}}\) optimization, we evaluate our algorithm without Q-loss optimization, which we will refer to as \(\alg^{-Q}\) for simplicity.


\textbf{RAD-s and RAD-m} \cite{rashid2018qmix} RAD is a data augmentation method that multiplies the uniform random variable $z \sim \mathcal{U}[\alpha, \beta]$ to the given state $s$, \textit{i.e.}, $s' = s * z$. Here, we set $\alpha$ and $\beta$ to be 0.8 and 1.2 respectively. If the random variable $z$ is single-variate variable, we denote it as RAD-s, otherwise (multi-variate variable), we denote it as RAD-m.

\subsection{Datasets and environments}
\label{subsec:dataset}
We select the simulator StarCraft Multi-Agent Challenges (SMAC) \cite{samvelyan2019starcraft} for evaluation of our proposed augmentation methods which is the most popular environment for the reason of complexity in cooperativeness and environments and the pre-collected dataset is from the open-source website \cite{formanek2023offthegrid}. For the assumption that the pre-collected dataset is small, we deliberately downsample the original dataset to be 3\% amount of the pre-collected dataset. We denote the downsampled dataset as \textit{original} dataset. 

SMAC consists of allies and enemies, where each team must kill others to win the battle. Allies receive the episode's total reward of 20 when they win a battle, as well as small rewards of 0.05 for killing an enemy and a payout equal to the amount of damage they dealt to adversaries. To win a battle, agents must cooperate among themselves to manage their group behavior, like \textit{focusing fire} while not overkilling the enemies, or \textit{kiting} to lure the enemies and kill them one by one. We report the results of scenarios \textit{3m vs 3m, 5m vs 6m, 8m vs 8m, 2s3z vs 2s3z and 3s5z vs 3s6z} with sup-optimal policies with return values in an episode.

\subsection{Diffusion model for \alg}
In our experiments, we selected the Denoising Diffusion Probabilistic Model (DDPM) detailed by \citet{ho2020denoising}, which incorporates a \texttt{Conv1D}-layer, to implement \alg approach. We adhere to the established DDPM hyperparameters for training and sampling, ensuring consistency across our processes.

The choice of DDPM is particularly suitable for our purposes due to its robustness and proven effectiveness in generating high-quality samples. However, it is important to note that \alg is versatile and not exclusively limited to DDPM or its variant DDIM. Our method is compatible with any diffusion model that can integrate a \texttt{Conv1D}-layer. This flexibility is advantageous because it allows us the possibility to experiment with different types of diffusion models, potentially enhancing our ability to fine-tune the algorithms according to specific requirements.


\subsection{Results} 
In our comprehensive experiments, we assess the performance of two variants of RL algorithms across a spectrum of datasets that have been enhanced by different augmentation methods. We report \alg augmented dataset performance tuned on the hyperparameter $\lambda$ among [0.5, 0.1, 0.01]. Our evaluation spans 10 pre-collected datasets, utilized in conjunction with 2 distinct RL algorithms, resulting in a total of 20 unique scenarios being tested. Among these, the \alg method demonstrates best performance in 14 out of the 20 tasks, marking a significant improvement in efficacy. Additionally, $\alg^{-Q}$, which does not utilize the maximization of the Q-function, achieves the best performance in 4 tasks, illustrating the impact of sophisticated augmentation techniques even without direct Q-function optimization. 
The RAD-s method excelled only in the \textit{3m vs 3m} scenario.

Overall, the results presented in Table \ref{table:main} underscore the significant advantages offered by our \alg method. Particularly, \alg has outperformed the baseline dataset by a substantial margin—17.3\% for datasets crafted under a medium policy datasets, and 12.9\% for those under a poor policy datasets, where the margin is calculated based on the normalized returns (normalized by maximum return 20). These results not only emphasize the effectiveness of \alg combined with original datasets but also highlight its superiority over prior augmentation methods.

Furthermore, it is intriguing to note that employing $\alg^{-Q}$ also results in an enhanced performance compared to both the original dataset and other state-based augmented datasets such as RAD-s and RAD-m. This observation is critical as it suggests that even without explicit optimization of the Q-values, the inherent capabilities of diffusion models can significantly elevate the quality of data augmentation, thereby boosting algorithm performance across various tasks. This is because the diffusion model generates new samples with high fidelity by accurately learning the true data distribution.

\paragraph{Dataset where \alg does not offer benefits.} 
In our examination, we observed that when utilizing datasets characterized by poor behavioral policies, our algorithm \alg exhibits performance levels that are similar to those achieved by the $\alg^{-Q}$ method. This is in stark contrast to the results obtained from datasets with medium behavioral policies, where \alg distinctly outperforms other approaches. This discrepancy in performance can be attributed to the inherent quality of the datasets being used.

\alg is designed to maximize the function $Q_{f_{\theta}(\tau_{k}, k)}(\textbf{o}, \textbf{a})$ during each epoch, with the upper bound of the maximum $Q$-value observed in the minibatch $\max_{\tau_{0} \in \mathcal{B}} Q_{\tau_{0}}(\textbf{o}, \textbf{a})$. This strategy is predicated on the assumption that there is potential for improvement in the dataset's quality through the application of our algorithm. However, if the initial quality of the dataset is exceedingly poor, the ceiling for enhancement is significantly lowered. Under such conditions, even with a considerable increase in the number of generated episodes, the improvement in dataset quality remains marginal compared to $\alg^{-Q}$.
\section{Conclusion and Future Work}
\label{sec:conclusion}
In this study, we use a diffusion model as an episodes augmentation module for offline MARL that mitigates the data scarcity problem. Specifically, we guide the diffusion model by the implicit Q-loss to generate multi-agent's cooperative actions, which is found to be effectively deploying a practical, and cooperativeness-oriented model. 

However, there is still room for improvement in \alg. First, we need to validate our algorithms across a broader range of observation and action spaces, such as those involving continuous actions. Second, we must explore ways to enhance dataset quality, especially when dealing with extremely poor behavioral policies. Nevertheless, our approach will work with any generative models, which we believe could be a potential future work by applying to the next generation models like Flow Matching.

\section*{Acknowledgements}
This work was supported by Center for Applied Research in Artificial Intelligence (CARAI) grant funded by Defense Acquisition Program Administration (DAPA) and Agency for Defense Development (ADD) (UD230017TD).

\bibliography{main}
\bibliographystyle{icml2024}

\end{document}